%% file: sample-journal.tex
\documentclass[acmtog,timestamp]{acmart}
\setcitestyle{square}

\usepackage{booktabs} 

\usepackage[ruled]{algorithm2e} 
\SetAlFnt{\small}
\SetAlCapFnt{\small}
\SetAlCapNameFnt{\small}
\SetAlCapHSkip{0pt}
\IncMargin{-\parindent}

\renewcommand\footnotetextcopyrightpermission[1]{} 
\acmDOI{0000001.0000001_2}
\usepackage{mwe}

\begin{document}
\title{High Resolution Face Completion with Multiple Controllable Attributes via Fully End-to-End Progressive Generative Adversarial Networks}


\author{Zeyuan Chen}
\affiliation{%
  \institution{North Carolina State University}
  }

\author{Shaoliang Nie}
\affiliation{%
  \institution{North Carolina State University}
  }
\author{Tianfu Wu}
\affiliation{%
  \institution{North Carolina State University}
  }
\author{Christopher G. Healey}
\affiliation{%
  \institution{North Carolina State University}
  }

\renewcommand\shortauthors{}

\begin{abstract}
We present a deep learning approach for high resolution face completion with multiple controllable attributes (e.g., male and smiling) under arbitrary masks.  Face completion entails understanding both structural meaningfulness and appearance consistency locally and globally to fill in ``holes" whose content do not appear elsewhere in an input image. It is a challenging task with the difficulty level increasing significantly with respect to high resolution, the complexity of ``holes" and the controllable attributes of filled-in fragments. Our system addresses the challenges by learning a fully end-to-end framework that trains generative adversarial networks (GANs) progressively from low resolution to  high resolution with conditional vectors encoding controllable attributes.

We design novel network architectures to exploit information across multiple scales effectively and efficiently. We introduce new loss functions encouraging sharp completion. We show that our system can complete faces with large structural and appearance variations using a single feed-forward pass of computation with mean inference time of 0.007 seconds for images at $1024\times 1024$ resolution. We also perform a pilot human study that shows our approach outperforms state-of-the-art face completion methods in terms of rank analysis. The code will be released upon publication. 
\end{abstract}

%
%


%
%


\begin{teaserfigure}
  \begin{center}
      \centering
      \begin{minipage}{1.0\textwidth}
          \centering
          \includegraphics[width=1.0\textwidth]{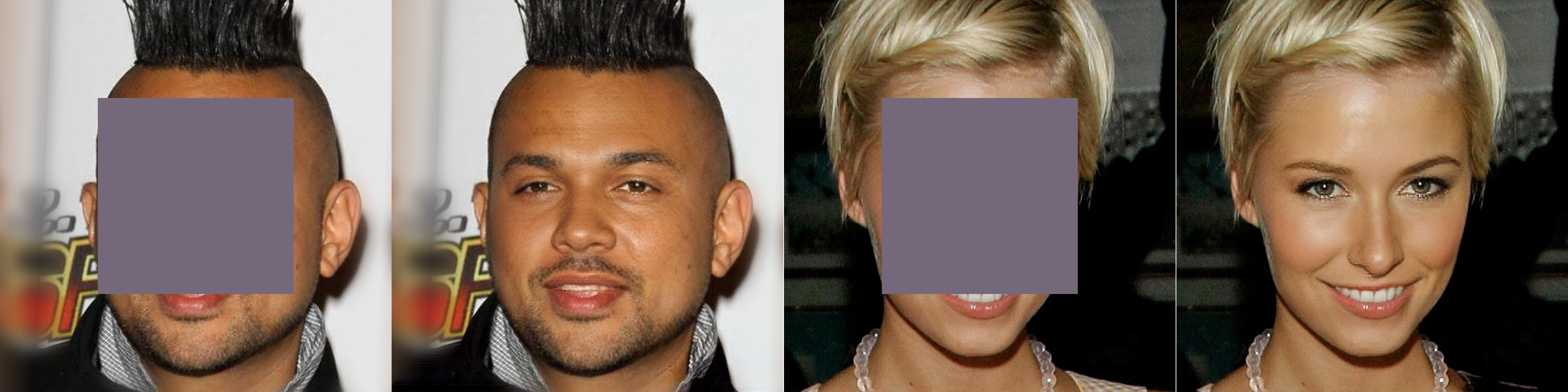} 
      \end{minipage}\vfill
      \begin{minipage}{1.0\textwidth}
          \centering
          \includegraphics[width=1.0\textwidth]{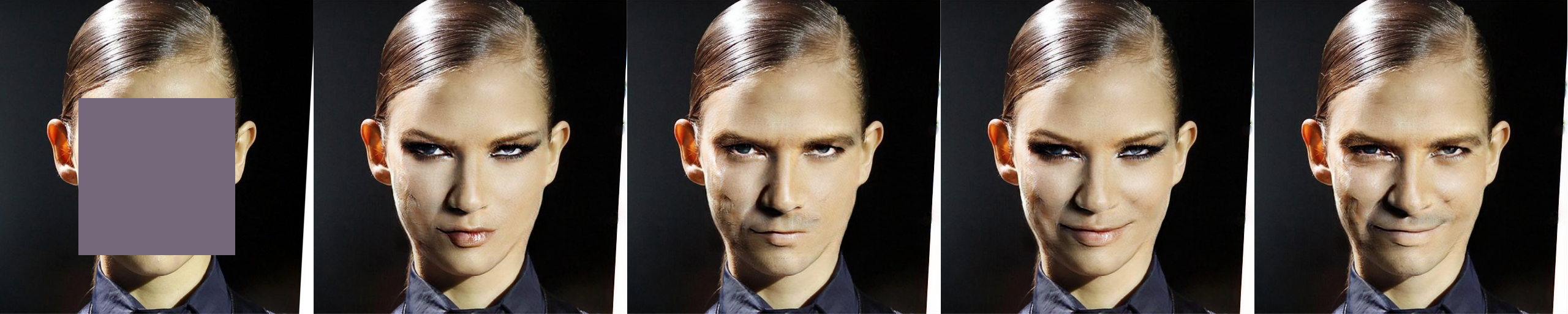} 
      \end{minipage}
  \end{center}%
  \caption{Face completion results of our method on CelebA-HQ~\cite{karras2017progressive}. Images in the left most column of each group are masked with gray color, while the rest are synthesized faces. \textit{Top}: our approach can complete face images at high resolution ($1024\times1024$). \textit{Bottom}: the attributes of completed faces can be controlled by conditional vectors. Attributes [``Male'', ``Smiling''] are used in this example. The conditional vectors of column two to five are [0, 0], [1, 0], [0, 1], and [1, 1] in which ``1'' denotes the generated images have the particular attribute while ``0'' denotes not. Images are at $512\times512$ resolution. All images best viewed enlarged.}
  \label{fig:teaser}
\end{teaserfigure}

\maketitle


\input{samplebody-journals}

\end{document}

%% file: samplebody-journals.tex
\section{Introduction}
Making things complete and more satisfying is always fascinating people in many creative ways. Image completion is a technique to replace target regions, either missing or unwanted, of images with synthetic content so that the completed images look natural, realistic and  appealing. The capability of seeing the unseen or realizing imagination has broad applications in visual content editing. 
Image completion can be divided in two categories: generic scene image completion and specific object image completion (e.g., human faces). Due to the well-known compositionality and reusability of visual patterns~\cite{geman2002composition}, target regions in the former usually have a high chance of finding similar patterns in either the surrounding context of the same image or images in an external image dataset subject to the context. Target regions in the latter are more specific, especially when large portions of essential
parts of an object are missing (e.g., facial parts in Figure~\ref{fig:teaser}). So, the completion entails  fine-grained understanding of the semantics, structures and appearance of images, and thus is a more challenging task. Face images have become one of the most popular source of images collected in people's daily lives and transmitted on social networks. We focus on human face completion in this paper.  

Two broad frameworks have been proposed in the literature of image completion: data similarity driven methods and data distribution based generative methods. In the first paradigm, texture synthesis or patch matching are usually used~\cite{efros1999texture,kwatra2003graphcut,criminisi2003object,wilczkowiak2005hole,komodakis2006image,barnes2009patchmatch,darabi2012image,huang2014image,wexler2007space}. Textures or patches are generated by finding similar exemplars in the known contexts and then stitched together to fill in the ``holes''. An alternative is the data-driven method~\cite{hays2007scene}, which searches a large image database for plausible patches based on context similarity. These methods are often utilized for generic scene image completion and their limitations are obvious. They are bound to fail when no similar exemplars can be found in either the context or the external dataset, and thus are not applicable to face completion (as well as other objects) as pointed out in ~\cite{iizuka2017globally,yeh2017semantic}. Instead of seeking similar exemplars, the second paradigm learns the underlying distribution governing the data generation with respect to the context. Much progress~\cite{li2017generative,yeh2017semantic,yang2016high,iizuka2017globally,denton2016semi,pathak2016context} has been made since the recent resurgence of deep convolutonal neural networks (CNNs)~\cite{cnn,AlexNet}, especially the generative adversarial network (GAN)~\cite{goodfellow2014generative}. 

We adopt the data distribution based generative method in this paper and address three important issues. \textit{First}, previous methods are only able to complete faces at low resolutions (e.g. $128\times128$~\cite{li2017generative} and $176\times216$~\cite{iizuka2017globally}). \textit{Second}, most approaches cannot control the attributes of the synthesized content. Previous works focus on generating realistic content. However, users may want to complete the missing parts with certain properties (e.g. smiling or not). \textit{Third}, most existing approaches require post processing or complex inference process. Generally, these methods~\cite{iizuka2017globally,li2017generative,yeh2017semantic} synthesize relatively low quality images from which the corresponding contents are cut and blended (e.g. with Poisson Blending~\cite{perez2003poisson}) with the original contexts. In order to complete one image, other approaches~\cite{yeh2017semantic,yang2016high} need run thousands of optimization iterations or feed an incomplete image to CNNs repeatedly at multiple scales. 

To overcome the above limitations, we propose a novel fully end-to-end progressive GAN to complete face images in high-resolution with multiple controllable attributes (see Figure~\ref{fig:teaser}). Our network is able to complete masked faces with high quality in a single forward pass without any post processing. It consists of two sub-networks: a completion network and a discriminator. Given face images with missing contents, the completion network tries to synthesize completed images that are indistinguishable from uncorrupted real faces, while keeping their contexts unchanged. The discriminator is trained simultaneously with the completion network to distinguish completed ``fake'' faces from real ones. Unlike most existing works~\cite{li2017generative,iizuka2017globally,yang2016high,denton2016semi} that use the Encoder-Decoder structures, we introduce a new architecture based on the U-Net~\cite{ronneberger2015u} that better integrates information across all scales to generate higher quality images. Moreover, we design new loss functions inducing the network to blend the synthesized content with the contexts in a realistic way. We adopt the training methodology of growing GANs progressively~\cite{karras2017progressive} to generate high-resolution images. A conditional version of our network is also designed so that \textit{N} attributes of the synthesized faces can be controlled by ~\textit{N}-dimensional vectors (Figure~\ref{fig:teaser}). We compared our method with state-of-the-art approaches on a high-resolution face dataset CelebA-HQ~\cite{karras2017progressive}. The results of both qualitative evaluation and a pilot user study showed that our approach completed face images significantly more naturally than existing methods, with improved efficiency.

The main contributions of this paper are:
\begin{itemize}
  \item We propose a novel approach that consolidates information across all scales to complete face images with arbitrary masks in much higher resolution than existing methods. 
  \item We further design a conditional version of our architecture to control multiple attributes of the synthesized content.
  \item Our framework is able to complete images in a single forward pass, without any post-processing.
\end{itemize}

\begin{figure*}
      \centering
      \begin{minipage}{1.0\textwidth}
          \centering
          \includegraphics[width=1.0\textwidth]{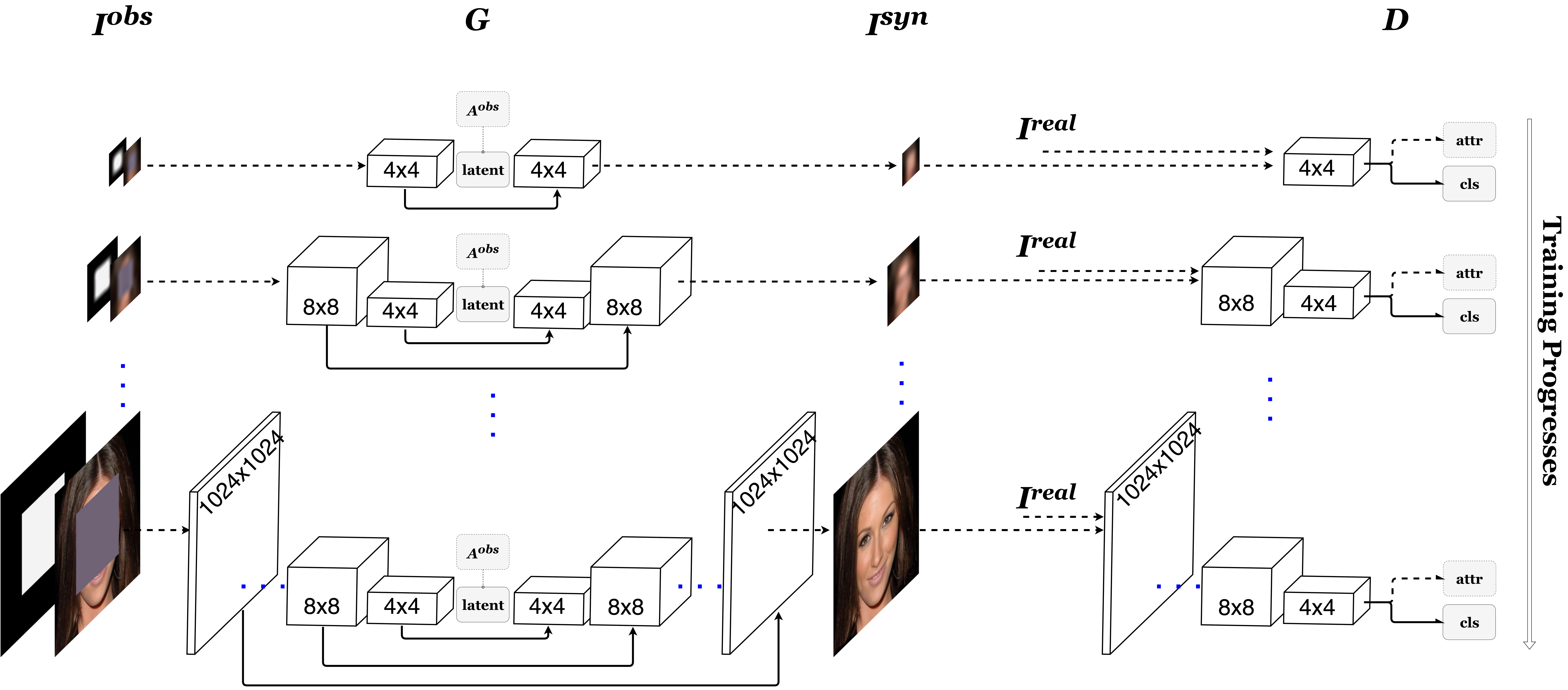}
      \end{minipage}\hfill
	 \caption{The overall architecture and training process of our approach. The training of the completion network (or the ``generator'' $G$) and the discriminator $D$ starts at low resolution ($4\times4$). Higher layers are added to both $G$ and $D$ progressively to increase the resolution of the synthesized images. The \framebox{$r$ x $r$} cubes in the figure represent convolutional layers that handle resolution $r$. For the conditional version, attribute labels $A^{obs}$ are concatenated to the latent vectors. The discriminator $D$ splits into two branches in the final layers: $D_{cls}$ that classifies if an input image is real, and $D_{attr}$ that predicts attribute vectors.}
     \label{fig:structure}
\end{figure*}%

\section{Related Work}
\subsection{Image Generation}
Generative models have been studied extensively to synthesize realistic and novel images by learning high-dimensional data distributions. Current work falls into three groups: variational autoencoders (VAE)~\cite{kingma2013auto,kingma2016improved}, autoregressive models~\cite{oord2016pixel,van2016conditional} and GAN~\cite{goodfellow2014generative}. VAE is easy to train, but the generated images are often blurry. Autoregressive models can synthesize sharp images, but they lack latent representations, which makes it more difficult to control the attributes of generated images than VAE or GAN. Additionally, the evaluation of autoregressive models is slow. GAN is able to generate sharp images from latent vectors (e.g. a 100-dimensional noise vector), whose architecture typically consists of two networks: a generator and a discriminator. The generator learns to synthesizes images that are indistinguishable from the training data while the discriminator is trained to differentiate between the generated images and real ones. The generator and discriminator are trained simultaneously and thus the training process is usually unstable. 

Many methods try to stabilize GAN and generate high quality images by designing new objective functions and architectures. The Deep Convolutional GAN (DCGAN)~\cite{radford2015unsupervised} produces images using a set of fractional-strided convolutions. The Laplacian GAN~\cite{denton2015deep} uses a Laplacian pyramid to generate images from coarse to fine by adding high frequency information at different layers. Unfortunately, these techniques are unable to synthesize high-resolution images. The main challenge is that, in higher resolutions, the discriminator can tell the differences between real and fake images more accurately, which causes a vanishing gradient problem. Recently, Xiang et al.~\cite{xiang2017effects} proposes a weight normalization approach and achieves better training performance. Karras et al.~\cite{karras2017progressive} put forward a progressive training mechanism to grow both the generator and discriminator from low to high resolution, and are able to generate realistic $1024\times1024$ images. The advantage of this methodology is that the networks do not have to handle information across all image scales at the same time, and instead can learn the holistic image structures first and then focus on producing finer details progressively. However, GANs cannot be applied to the image completion task directly because they aim at generating random natural images, but are not constrained by the image context.

\subsection{Image Completion}
There is a large body of image completion literature. Early non-learning based algorithms~\cite{efros1999texture,bertalmio2000image,bertalmio2003simultaneous} complete missing content by propagating information from known neighborhoods, based on low level cues or global statistics~\cite{levin2003learning}. Texture synthesis and patch matching based approaches~\cite{efros1999texture,kwatra2003graphcut,criminisi2003object,wilczkowiak2005hole,komodakis2006image,barnes2009patchmatch,darabi2012image,huang2014image,wexler2007space} find similar structures from the context of the input image or from an external database~\cite{hays2007scene} and then paste them to fill in the holes. These methods assume that similar textures or patches of the missing contents can be found in the known regions. Recent work~\cite{pathak2016context,iizuka2017globally,yeh2017semantic} has compared learning based methods with aforementioned approaches to produce large missing content. The results show that non-learning based approaches often synthesize content that is inconsistent with the global structures (e.g. using \textit{mouths} to fill in holes at \textit{eye} locations) while the learning based models can produce reasonable results. 

Many researchers focus on the face completion problem. The Graph Laplace method~\cite{deng2011graph} uses a spectral-graph-based algorithm to repair occluded face images. The Visio-lization~\cite{mohammed2009visio} completes faces with realistic and variant characteristics using a combination of global and local models. However, these approaches can handle only low-resolution images with limited shapes of masks. 

Recent learning based methods have shown the capability of CNNs to complete large missing content. The completion models are different from the generative models (e.g. GANs): the former need to complete corrupted images with plausible content while the latter focus on generating completely fake, yet realistic images from latent vectors. Based on existing GANs, the Context Encoder (CE) ~\cite{pathak2016context} encodes the contexts of masked images to latent representations, and then decodes them to natural content images, which are pasted into the original contexts for completion. However, the images generated by CE are often blurry and have inconsistent boundaries along the seams between content and context. Given a trained generative model, Yeh et al.~\cite{yeh2017semantic} proposed a framework to find the most plausible latent representations of contexts to complete masked images. But this work depends heavily on the quality of the pre-trained generative models.  The Generative Face Completion model (GFC)~\cite{li2017generative} and the Global and Local Consistent model (GL)~\cite{iizuka2017globally} use both global and local discriminators, combined with post processing, to complete images more coherently. Though GFC and GL models are trained with random rectangular masks, they can handle masks with arbitrary shapes as well. Unfortunately, these two approaches can only complete face images in relatively low resolutions (e.g. 176 x 216~\cite{iizuka2017globally}). Yang et al.~\cite{yang2016high} combined a global content network and a texture network, and trained networks at multiple scales repeatedly to complete high-resolution images ($512\times512$). Like the patch matching based approaches, Yang et al. assume that the missing content always shares some similar textures with the context, which is improbable for the face completion task. 

\subsection{Generative Models with Controllable Attributes}
There are many researchers studying how to control the attributes of synthesized images from generative models. The Conditional GAN (CGAN)~\cite{mirza2014conditional} trains the networks conditioned on attribute vectors (e.g. one-hot class labels) that are used to control properties of produced images explicitly during evaluation. For instance, the trained model of CGAN on MNIST~\cite{lecun1998mnist} can generate images of digits zero to nine depending on a label vector. Kaneko et al.~\cite{kaneko2017generative} extend this work and design multi-dimensional controllers to manipulate the properties (e.g. young or old) of face images. Olszewski et al.~\cite{olszewski2017realistic} generates dynamic textures for a target face by referencing a source video sequence. Unlike the CGANs that make the generators or discriminators conditioned on the attribute code directly, the information model (InfoGAN)~\cite{chen2016infogan} is able to control continuous (e.g. width of digit) and discrete (e.g. category) attributes of images by preserving the attribute information during the generation process. In additional to distinguishing real images from fake ones, InfoGAN uses auxiliary networks to check whether the latent information predicted from the generated images is close to the input latent code. The Categorical GAN (CatGAN)~\cite{springenberg2015unsupervised} controls the categories of generated images based on the assumption that real images have peaked class distributions while fake images should be uniformly distributed. Salimans et al.~\cite{salimans2016improved} change the discriminator to a ``\textit{K+1}'' classifier by adding a ``generated'' class to the original \textit{K} image classes. Their model is able to generate images of multiple classes with one generative model. Recent image-to-image translation networks are able to transfer images to other domains by explicitly~\cite{choi2017stargan} or implicitly~\cite{zhu2017unpaired,isola2016image} learning image characteristics. However, there are few completion models that are able to manipulate the properties of synthesized contents. Our conditional completion model is built on these generative approaches and can complete corrupted images with multiple controllable attributes.

\begin{figure*}
      \centering
      \begin{minipage}{1.0\textwidth}
          \centering
          \includegraphics[width=1.0\textwidth]{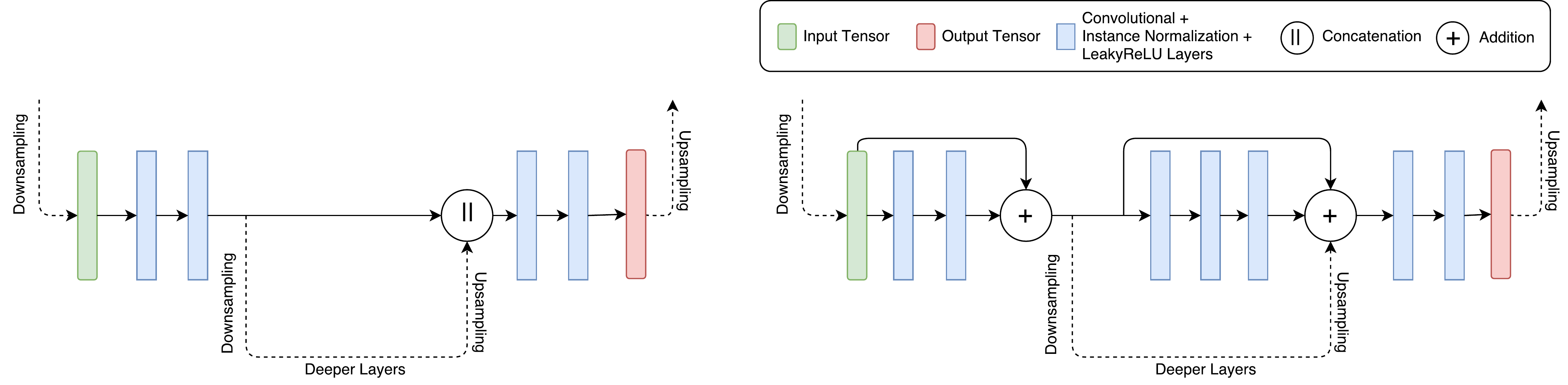} %
      \end{minipage}\hfill
	 \caption{Illustrations of a single layer of our architecture. There are skip connections between mirrored encoder and decoder layers. Left: the structure of the completion network; the skip connection is a copy-and-concatenate operation. This structure helps preserve the identity information between the synthesized images and real faces, resulting in little deformation. Right: the structure of the conditional completion network; residual connections are added to the encoder, and the skip connections are residual blocks instead of direct concatenation. The attributes of the synthesized contents can be manipulated more easily with this structure. Each blue rectangle represents a set of Convolutional, Instance Normalization and Leaky Rectified Linear Unit (LeakyReLU)~\cite{maas2013rectifier} layers.}
     \label{fig:layer}
\end{figure*}%

\section{Approach}
In this section, we first formulate the problem of image completion. Then, we present details of the proposed fully end-to-end progressive generative adversarial network. The overall structure of our networks is shown in Figure~\ref{fig:structure}.

\subsection{Problem Formulation}
Denote by $\Lambda$ an image lattice (e.g., $1024\times 1024$ pixels). Let $I_{\Lambda}$ be an RGB image defined on the lattice $\Lambda$. Denote by $\Lambda_t$ and $\Lambda_c$ the target region to complete and the remaining context region respectively which form a partition of the lattice, i.e., $\Lambda_t\cap\Lambda_c=\emptyset$ and $\Lambda_t\cup\Lambda_c=\Lambda$. Without loss of generality, we assume $\Lambda_t$ is a single connected component region (e.g., a rectangular mask in Figure~\ref{fig:teaser}). $I_{\Lambda_t}$ is masked out with the same gray pixel value. Let $M_{\Lambda}$ be a binary mask image with all pixels in  $M_{\Lambda_t}$ being $1$ and all pixels in  $M_{\Lambda_c}$ being $0$. For notational simplicity, we will omit the subscripts $\Lambda$, $\Lambda_t$ and $\Lambda_c$ when the text context is clear. 

The objective of image completion is to generate a synthesized image $I^{syn}$ that looks natural, realistic and appealing for an observed image $I^{obs}$ with the target region $I^{obs}_{\Lambda_t}$ masked out. Furthermore, the generator can be controlled  with respect to a set of attributes which are assumed to be independent from each other. Let $A=(a_1, \cdots, a_N)$ be a $N$-dim vector with $a_i\in\{0,1\}$ encoding if a corresponding attribute appears ($a_i=1$) or not ($a_i=0$) such as the male and smiling attributes in Figure~\ref{fig:teaser}.  We define the generator as,  
\begin{equation}
I^{syn} = G(I^{obs}, M, A; \theta_G)\label{eqn:G}
\end{equation}
where $\theta_G$ collects all parameters of the generator (to be elaborated later), and the context regions, $I^{syn}_{\Lambda_c}$ and $I^{obs}_{\Lambda_c}$, are kept very similar. 

Following the data distribution based generative methods, the generator needs to tightly approximate the underlying conditional probability model $p_G(I^{syn}|I^{obs}, M, A)$ in the high dimensional huge image space, and to implement a one-pass sampler which can generate a typical sample from the probability model (Eqn.~\ref{eqn:G}), thus learning the generator $G(\cdot)$ is an extremely difficult task. 

\subsection{The Proposed Fully End-to-End Progressive GAN}
Thanks to the recently proposed Generative Adversarial Networks (GAN)~\cite{goodfellow2014generative}, a generator that learns data distribution and computes typical sample can be trained under a minimax game setting. Denote by $p_G(I)$ and $p_{data}(I)$ the generator distribution and the data distribution respectively, and the former is trained to match the latter. GAN parametrizes $p_G(I)$ using a generator network $G$ which transforms a noise random variable (e.g., white noise) $z$ into a sample $G(z)$, overcoming the challenges of trying to compute probability to every $I$ in the data distribution in an explicit way. Under the minimax game, an adversarial discriminator network $D$ is trained simultaneously which aims to tell the generated sample $G(z)$ apart from the real data based on binary classification between real and fake. For a given generator $G$, the optimal discriminator is $D(I)=\frac{p_{data}(I)}{p_{data}(I)+p_{G}(I)}$ under the Nash equilibrium. In the work of Goodfellow et al.~\cite{goodfellow2014generative}, the minimax game is formulated by, 
\begin{align}
\nonumber \min_{G} \max_{D} \mathcal{L}_{adv}(G,D) = & E_{z\sim p_{noise}(z)}[1-\log{(1-D(G(z)))}] + \\
& E_{I\sim p_{data}(I)}[\log{D(I)}] \label{eqn:GAN}
\end{align}    
where $\mathcal{L}_{adv}(G,D)$ is the adversarial loss function and $E[\cdot]$ represents the expectation. The proposed method is built on the GAN.

\textit{In our model, the generator} takes as input the observed corrupted image, the binary mask image and the attribute vector and outputs a completed image. It consists of two components, 
\begin{equation}
G(I^{obs}, M, A;\theta_G) = G_{compl}(G_{enc}(I^{obs}, M;\theta_G^{enc}), A; \theta_G^{compl})
\end{equation}
where $G_{enc}(\cdot)$ encodes an input pair $(I^{obs}, M)$ to a latent low dimensional vector. The latent vector is concatenated with the attribute vector. The concatenated vector plays the role of the noise random variable $z$ in the original GAN. Then, $G_{compl}(\cdot)$ transforms the concatenated vector to a sample (i.e., the completed image). $G_{enc}$ and $G_{compl}$ are mirrored to each other.  $\theta_G=(\theta_G^{enc}, \theta_G^{compl})$. 

\textit{In our model, the discriminator} takes as input either the ground-truth uncorrupted image or the completed image from the generator and has two output branches. It consists of three components: a shared feature backbone and two head classifiers. We have, 
\begin{equation}
D(I;\theta_D) = \{D_{cls}(F(I;\theta_D^F); \theta_D^{cls}), D_{attr}(F(I;\theta_D^0); \theta_D^{attr})\}
\end{equation}
Where the feature backbone,  $F(I;\theta_D^0)$  computes the feature map for an input image. On top of the feature map, the first head classifier, $D_{cls}(I; \theta_D^{cls})=D_{cls}(F(I;\theta_D^F); \theta_D^{cls})$ computes binary classification between real and fake, and the second one, $D_{attr}(I; \theta_D^{attr})=D_{attr}(F(I;\theta_D^F); \theta_D^{attr})$ predicts an attribute vector. All the parameters of the discriminator are collected by $\theta_D=(\theta_D^F, \theta_D^{cls}, \theta_D^{attr})$. $\theta=(\theta_G, \theta_D)$ will be learned end-to-end. We note that the discriminator is only needed in training. We will omit the notations for parameters in equations when no confusion is caused. 

Next, we elaborate on the details of training which includes the generation of $I^{obs}$ and its attribute $A^{obs}$, the loss functions, the network architectures and the progressive training we propose for high resolution face completion.  

\subsubsection{Generating $I^{obs}$ and $A^{obs}$} Let $I^{real}_{\Lambda}$ and $A^{real}$ be an uncorrupted face image and its annotated attribute vector. To generate $I^{obs}_{\Lambda}$, we first sample a mask $M$, then use it to mask out the target regions. We use approaches similar to the one proposed by the context encoder method~\cite{pathak2016context}. First, starting with an all-zero one-channel image, a rectangular region of random size and location is chosen. Second, a low resolution noise (e.g. $4\times4$, drawn from uniform distribution) image is generated and then up-sampled to the size of the chosen rectangle with bi-linear interpolation. In this way, we can construct a rectangular region with continuous random values. Then the image is converted to a binary mask with thresholding.
Denote by $M\sim p_{mask}(M)$ a mask sample. Since we only know the occurrence of attributes in $I^{real}_{\Lambda}$, we can not infer how a mask $M$ affects the occurrence. To create the attribute vector $A^{obs}$ for $(I^{obs}_{\Lambda}, M)$, we define it as a fake attribute vector, 
\begin{equation}
	A^{obs}=\begin{cases}
    A^{real}, & \text{if $p<0.5$}.\\
    (a_1, \cdots, 1-a_i, a_{i+1}, \cdots, a_N); \forall j, a_j\in A^{real}, & \text{otherwise}.
    \end{cases}
\end{equation}
where $p\sim \textit{U}\,(0,\, 1)$ and $i \in [1, N]$ is a randomly chosen index. We will denote by $A^{obs}\sim p_{attr}(A^{real})$ an attribute vector sample. 

\subsubsection{Loss Functions} 
Beside extending the original adversarial loss function, we design three  new loss functions to enforce sharp image completion.

\paragraph{Adversarial Loss} Given an uncorrupted image $I^{real}$, its attribute vector $A^{real}$, a mask $M$ and the corresponding corrupted image $I^{obs}$, and a fake attribute vector $A^{obs}$, 
we define the loss by, 
\begin{align}
\nonumber l(I^{real}, M, I^{obs}, A^{obs} |G, D) =  &(1-\log{(1-D_{cls}(I^{syn})})+\\
& \log{D_{cls}(I^{real})} 
\end{align}
where $I^{syn}=G(I^{obs}, M, A^{obs}))$.
Similar to Eqn.~\ref{eqn:GAN}, we have the expected loss, 
\begin{align}
\mathcal{L}_{adv}(G,D) =  E_{\substack{I^{real}\sim p_{data}(I),\\ M\sim p_{mask}(M),\\A^{obs}\sim p_{attr}(A^{real})}}[l(I^{real}, M, I^{obs}, A^{obs} |G, D)]\label{eqn:expectedAdv}
\end{align}
In the following, we will omit definitions of the expected losses of different loss terms. 

\paragraph{Attribute Loss} For the attribute prediction head classifier in the discriminator, we define the attribute loss based on cross-entropy between the predicted attribute vector, $\hat{A}^{real}=D_{attr}(I^{real})$ and $\hat{A}^{obs}=D_{attr}(I^{obs})$ and the corresponding targets, $A^{real}$ and $A^{obs}$ for a real uncorrupted image and a synthesized image respectively. We have,
\begin{align}
\nonumber l_{attr}(&I^{real}, A^{real}, M, I^{obs}, A^{obs} |G, D)=\\
\nonumber &\sum_{i=1}^N (a_i^{real}\log \hat{a_i}^{real}+(1-a_i^{real})\log (1-\hat{a_i}^{real}))+\\
&\sum_{i=1}^N (a_i^{obs}\log \hat{a_i}^{obs}+(1-a_i^{obs})\log (1-\hat{a_i}^{obs})).\label{eqn:attr}
\end{align}

\paragraph{Reconstruction Loss}
Since our method generates the entire completed face rather than only the target region, we define a wighted reconstruction loss $l_{rec}$ to preserve both the target region and the context region, which is defined as, 
\begin{align}
\nonumber l_{rec}(I^{real}, M,  I^{obs}, & A^{obs}|G) = \|\alpha \cdot M \cdot (I^{real}-I^{syn}) \|_1+\\
&\|(1-\alpha) \cdot (1-M) \cdot (I^{real}-I^{syn}) \|_1. \label{eqn:rec}
\end{align}
where $\odot$ represents element-wise multiplication and $\alpha$ is the trade-off parameter.

\paragraph{Feature Loss}
In additional to the reconstruction loss in terms of pixel values, we also encourage the synthesized image to have a similar feature representation~\cite{johnson2016perceptual} based on a pre-trained deep neural network $\phi$. Let $\phi_{j}$ be the activations of the $jth$ layer of $\phi$, the feature loss is defined by
\begin{equation}
l_{feat}(I^{real}, M, I^{obs}, A^{obs}|\phi, G) = \| \phi_{j}(I^{real}) - \phi_{j}(I^{syn})) \|_{2}^{2}.\label{eqn:feat}
\end{equation}
In our experiments, $\phi_{j}$ is the $relu2\_2$ layer of a 16-layer VGG network~\cite{simonyan2014very} pre-trained on the ImageNet dataset~\cite{russakovsky2015imagenet}.

\paragraph{Boundary Loss}
To make the generator learn to blend the synthesized target region with the original context region seamlessly, we further define a close-up reconstruction loss along the boundary of the mask. Similar to~\cite{yeh2017semantic}, we first create a weighted kernel $w$ based on the mask image $M$. $w$ is computed by blurring the mask boundary in $M$ with a mean filter so that the pixels closer to the mask boundary are assigned larger weights. The kernel size of the mean filters is seven in our experiments.
\begin{equation}
l_{bdy}(I^{real}, M, I^{obs}, A^{obs}|G) = \| w \odot (I^{real} - I^{syn}) \|_{1}. \label{eqn:bdy}
\end{equation}

Our model is trained end-to-end by integrating Eqn.~\ref{eqn:expectedAdv} and the expected losses of Eqn.~\ref{eqn:attr}, Eqn.~\ref{eqn:rec}, Eqn.~\ref{eqn:feat} and Eqn.~\ref{eqn:bdy} under the minimax game setting. We have,
\begin{align}
\nonumber \min_G\max_D &\mathcal{L}(G,D) = \mathcal{L}_{adv}(G,D) + \lambda_{attr}\cdot \mathcal{L}_{attr}(G,D) + \\ &\lambda_{rec}\cdot \mathcal{L}_{rec}(G) + \lambda_{feat}\cdot \mathcal{L}_{feat}(G,\phi) + \lambda_{bdy}\cdot \mathcal{L}_{bdy}(G).
\end{align}
Where $\lambda_{\cdot}$'s are trade-off parameters between different loss terms.

\paragraph{Training without Multiple Controllable Attributes.} To that end, since it is a special case of the proposed formulation stated above, we can simply remove the components involving attributes such as the attribute loss in a straightforward way. The resulting system still enjoys end-to-end learning.   

\subsubsection{Network Architectures} As illustrated in Figure~\ref{fig:structure}, the generator $G$ in our model is implemented by a U-shape network architecture consisting of the first component $G_{enc}$ transforming the observed image and its mask to a latent vector and the second component $G_{compl}$ transforming the concatenated vector (latent and attribute) to the completed image. There are residual connections between layers in $G_{enc}$ and the counterpart in $G_{compl}$ similar in the spirit to the U-Net~\cite{ronneberger2015u} and the Hourglass network~\cite{newell2016stacked} to consolidate information across multiple scales. Figure~\ref{fig:layer} illustrates the two structures of a layer in the generator for training without and with attributes respectively, which are adapted from the U-Net and Hourglass network. \textit{Detailed specifications of the generator and the discriminator will be provided in the supplementary material.}   

\subsubsection{Progressive Training}
To address the challenge of stabilizing the training of GANs with faster convergence rates, we follow the very recent work on training GANs progressively~\cite{karras2017progressive}. It  starts with the lowest resolution (i.e. $4\times4$). After running a certain number of iterations, higher resolution layers are added to both the generator \textit{G} and discriminator \textit{D} at the same time. To avoid sudden changes to the trained parameters, the added layers are faded into the networks smoothly. All parameters are still trainable during the growing of networks. 

At a resolution lower than $1024\times1024$, the input face images, masks and real images are all down-sampled with average pooling to fit the given scale. The advantage of  progressive training  is that the networks can learn the structures of missing contents from coarse to fine incrementally. The holistic and local information are aggregated at multiple spatial scales. We note that since we do not need the global-and-local discriminator structure like previous works~\cite{iizuka2017globally,li2017generative}, the sizes and shapes of our masks could be arbitrary during the training, instead of being limited to rectangular regions with certain sizes. 

One of the major challenges of generating high resolution images is the limitation of Graphics Processing Unit (GPU) memory. Most completion networks use Batch Normalization~\cite{ioffe2015batch} to avoid covariate shift. However, with the limited GPU memory, only a small number of batch sizes are supported at high resolution, resulting in low quality of generated images. We use the Instance Normalization~\cite{ulyanov2016instance}, similar to Zhu et al.~\cite{zhu2017unpaired}, and update \textit{D} with a history of completed images instead of the latest generated one~\cite{shrivastava2016learning} to stabilize training.

\begin{figure*}
      \centering
      \begin{minipage}{1.0\textwidth}
          \centering
          \includegraphics[width=1.0\textwidth]{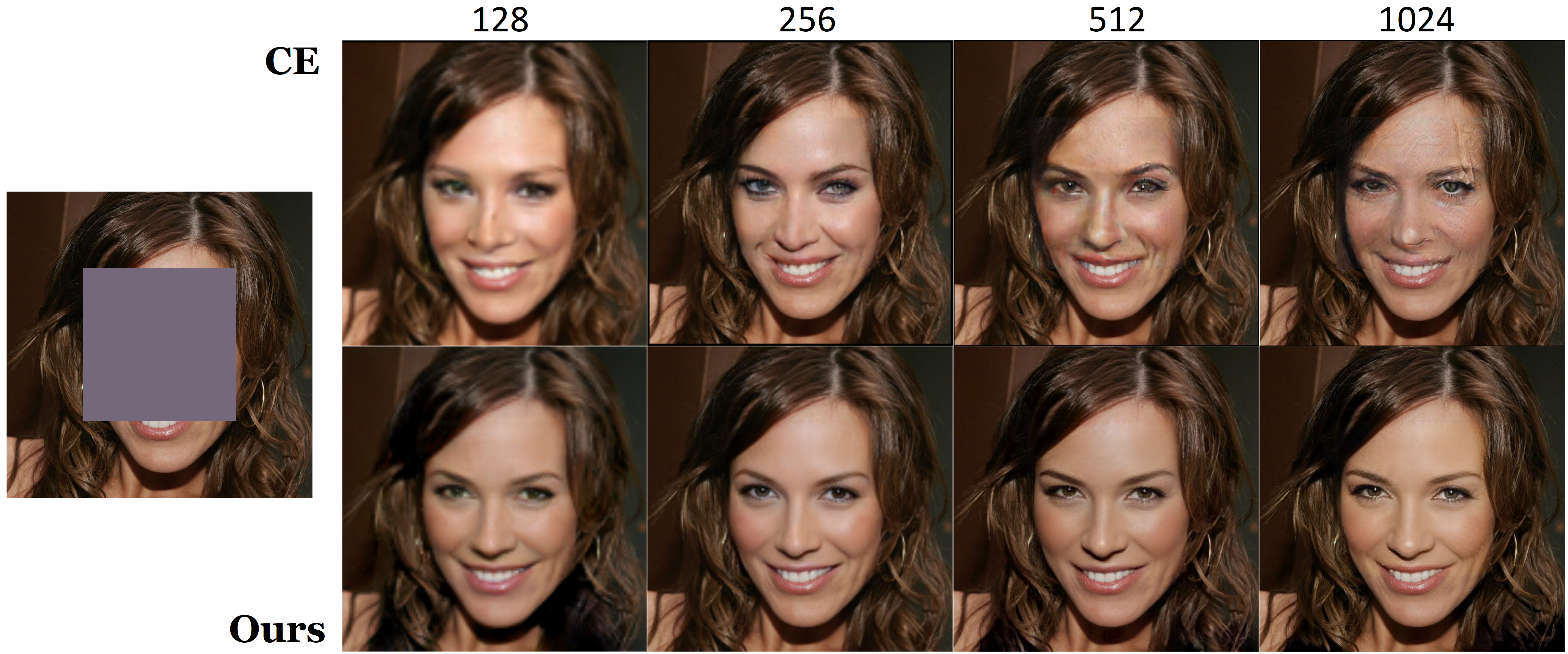} %
      \end{minipage}\hfill
	  \caption{Comparison with Context Encoder on high-resolution face completion. The top row are images generated by CE and the bottom row are our results. With increasing resolution (from $128\times128$ to $1024\times1024$), CE generated more distorted images while our method produced sharper faces with more details.}
     \label{fig:grow}
\end{figure*}%

\begin{figure*}
      \centering
      \begin{minipage}{0.33\textwidth}
          \centering
          \includegraphics[width=1.0\textwidth]{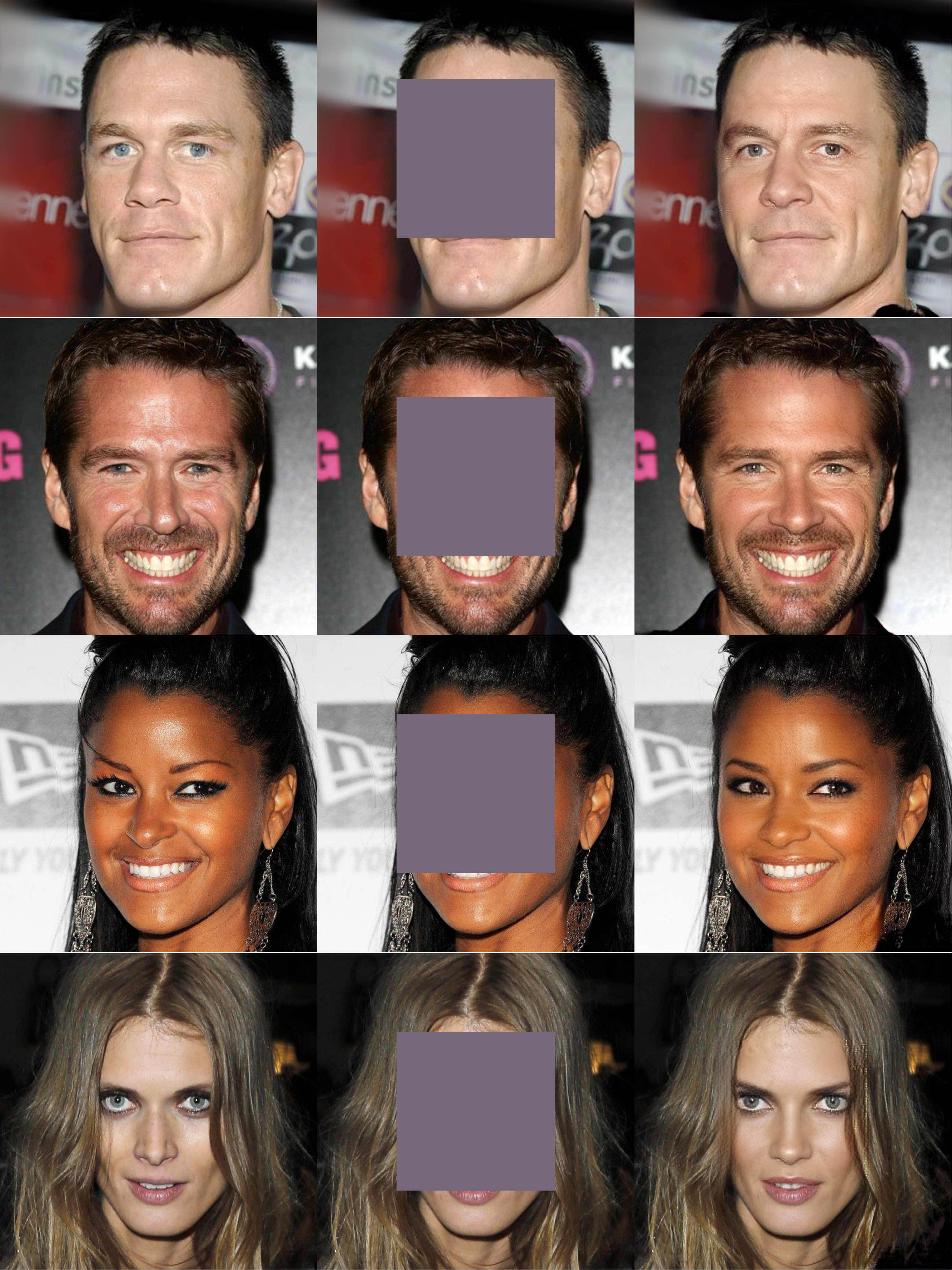} %
      \end{minipage}\hfill
      \begin{minipage}{0.33\textwidth}
          \centering
          \includegraphics[width=1.0\textwidth]{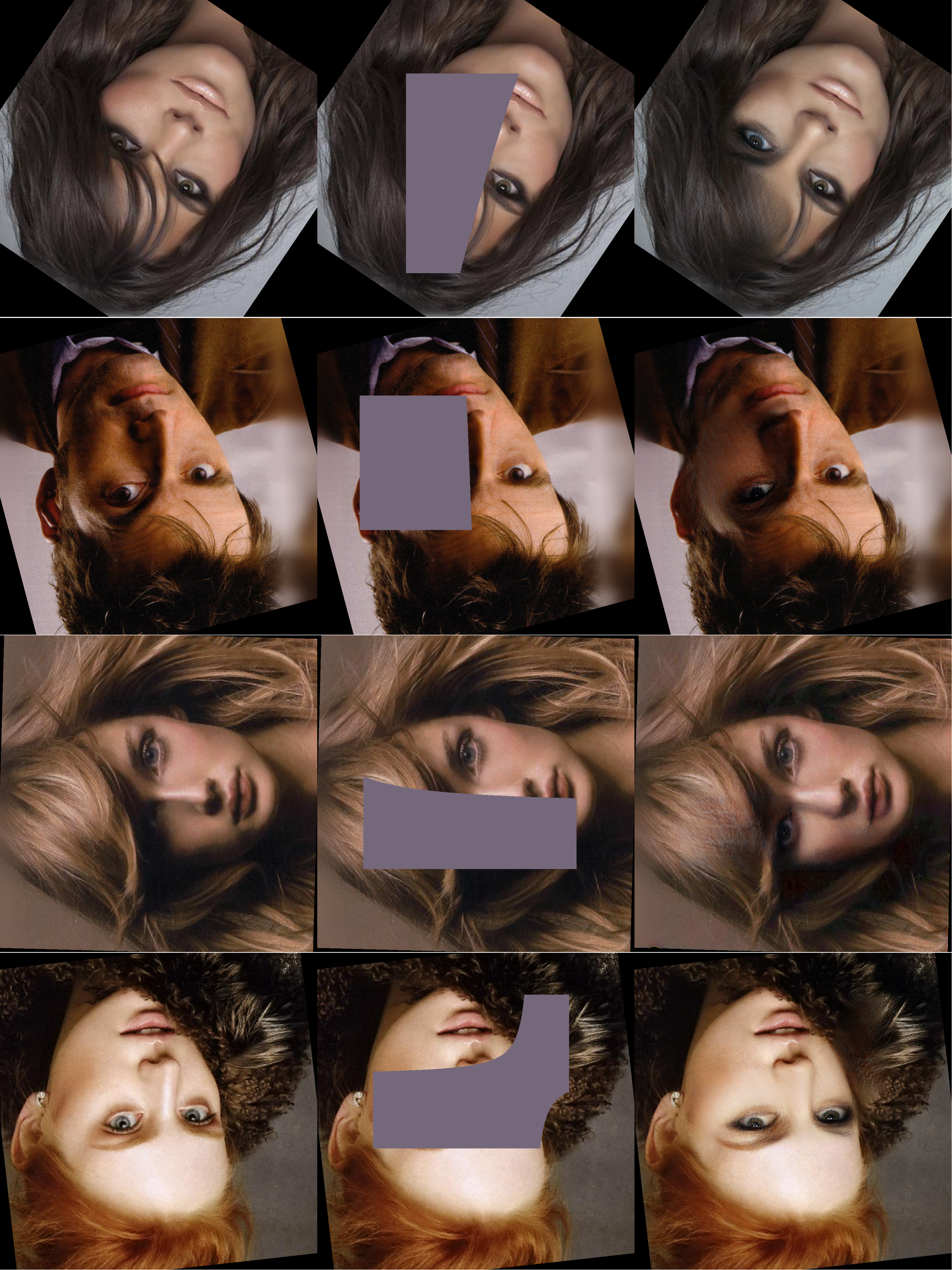} %
      \end{minipage}\hfill
      \begin{minipage}{0.33\textwidth}
          \centering
          \includegraphics[width=1.0\textwidth]{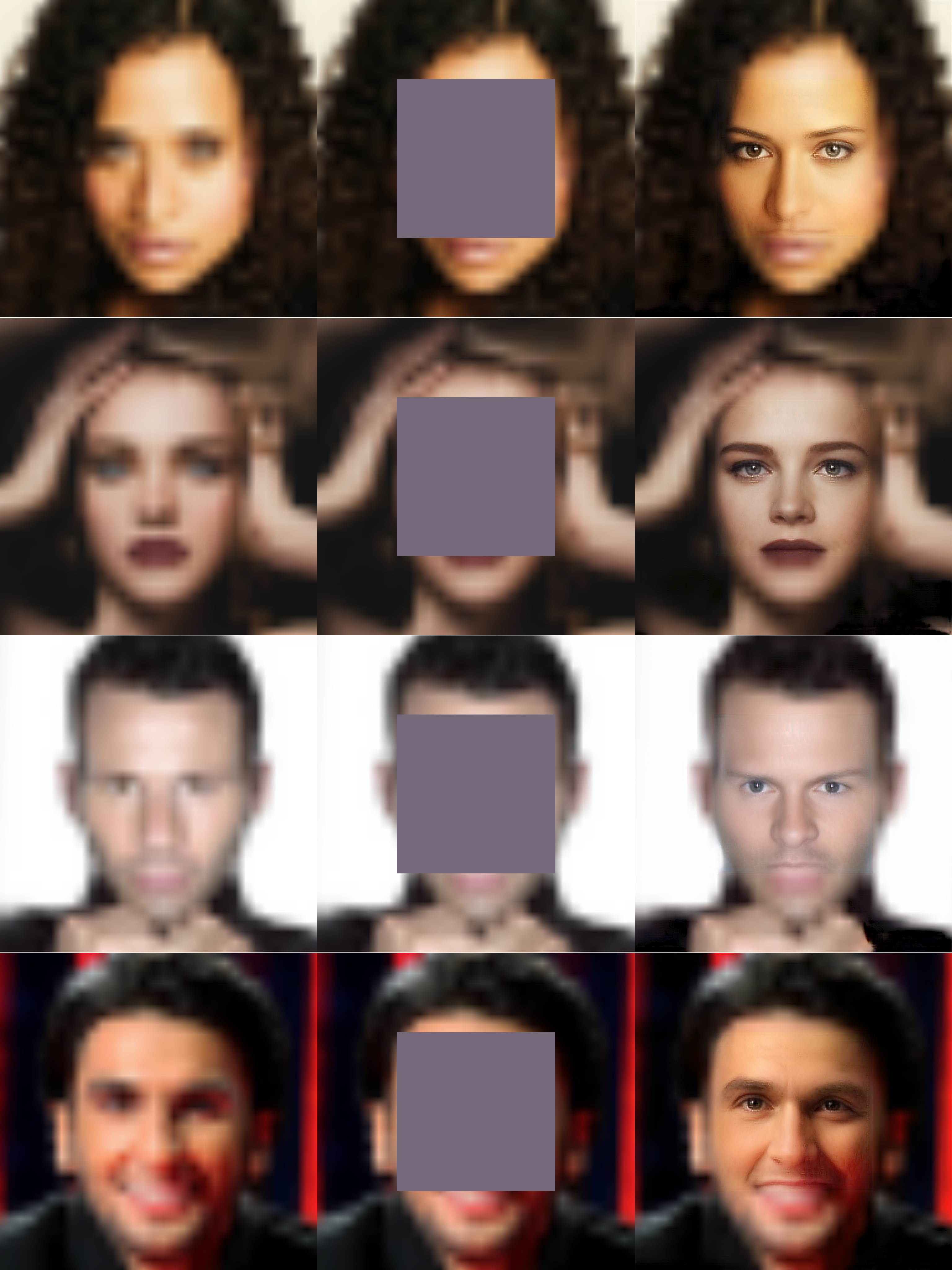} %
      \end{minipage}\hfill
	  \caption{Examples of high-resolution face completion results from our approach. All images are at $1024\times1024$ resolution. For each group, the left-most column are real images, the middle column are masked images and the right-most column are images synthesized by our model. Left group: images are masked with $512\times512$ holes in the center. Middle group: images are randomly flipped, rotated and covered by masks with arbitrary shapes and sizes. Right group: images with blurry contexts are completed by a model trained on clear contexts.}
     \label{fig:high}
\end{figure*}%

\begin{figure*}
      \centering
      \includegraphics[width=0.95\textwidth]{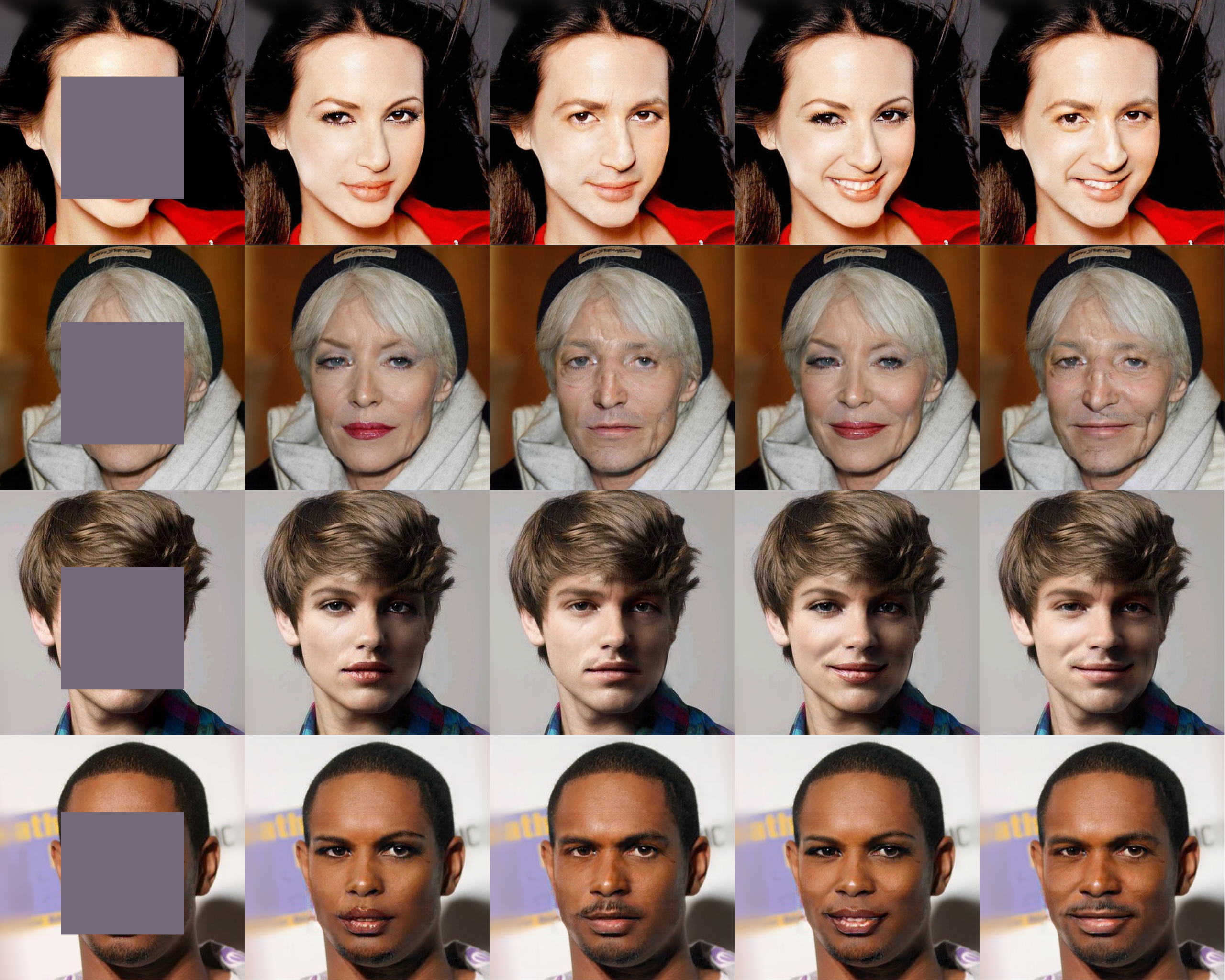} 
	  \caption{Examples of images generated by our conditional model. All images are at $512\times512$ resolution. The leftmost column are masked images, and the rest are generated faces. The characteristics of synthesized images can be controlled by attribute vectors explicitly. Attributes [``Male'', ``Smiling''] are used in this example. The attribute vectors of column two to five are [0,0], [1,0], [0,1], and [1,1] (``1'' denotes yes while ``0'' denotes no). Our model also learns to add detailed features to the context to make the face properties more consistent with the attribute labels, for instance adding chin beard to ``Male'' images in the second row, without affecting the holistic structures.}
      \label{fig:attr}
\end{figure*}

\section{Experiments}
In this section, we first demonstrate our models' ability to complete high-resolution face images in several challenging scenarios through experiments. Additionally, we show examples of controlling the attributes of synthesized faces. In the end, we compare our method with state-of-the-art approaches in low resolution with a pilot user study. 

\subsection{Datasets and Experiment Settings}
We used the CelebA-HQ~\citep{karras2017progressive} dataset for evaluation. It contains 30,000 aligned face images at $1024\times1024$ resolution. Similar to previous methods~\cite{yeh2017semantic}, we split the dataset randomly: 3,000 images for testing, and the remaining 27,000 for training. The CelebA-HQ was chosen over the original CelebA~\cite{liu2015deep} dataset not only because CelebA-HQ has higher resolution images, but also because it is a cleaner dataset with significantly fewer artifacts and more consistent quality. The images were scaled to $128\times128$ for the user study and $512\times512$ for the attribute controlling task. The remaining experiments used $1024\times1024$ images. There were two types of masks: center and random. The center mask was a square region in the middle of the image with a side length of half the size of the image. The random masks, which were generated similar to the method of Pathak et al.~\cite{pathak2016context}, were mostly continuous regions with arbitrary shapes, sizes and locations and covered about $10\%$ to $30\%$ of the original images. 

In the experiments, the reconstruction trade-off parameter was set to $\alpha=0.7$ to focus more on the target region. To balance the effects of different objective functions, we used  $\lambda_{attr}= 2$, $\lambda_{rec}=500$, $\lambda_{feat}=10$, and $\lambda_{bdy}=5000$. The Adam solver~\cite{kingma2014adam} was employed with a learning rate of 0.0001. 

\subsection{High-Resolution Face Completion}
\subsubsection{Comparison with the Context Encoder} 
Our method was compared with the Context Encoder (CE)~\cite{pathak2016context} on high-resolution face completion. Since the original networks of CE were designed for $128\times128$ images, we used a naive approach to fit it to different resolutions. One, two, and three convolutional layers were added to the encoder, decoder and discriminator for $256\times256$, $512\times512$ and $1024\times1024$ networks respectively. The result (Figure~\ref{fig:grow}) shows that, when the resolution increased, our method learned details incrementally and synthesized sharper faces, while CE generated poorer images with more distortions.

\begin{figure}
      \centering
      \begin{minipage}{0.42\textwidth}
          \centering
          \includegraphics[width=1.0\textwidth]{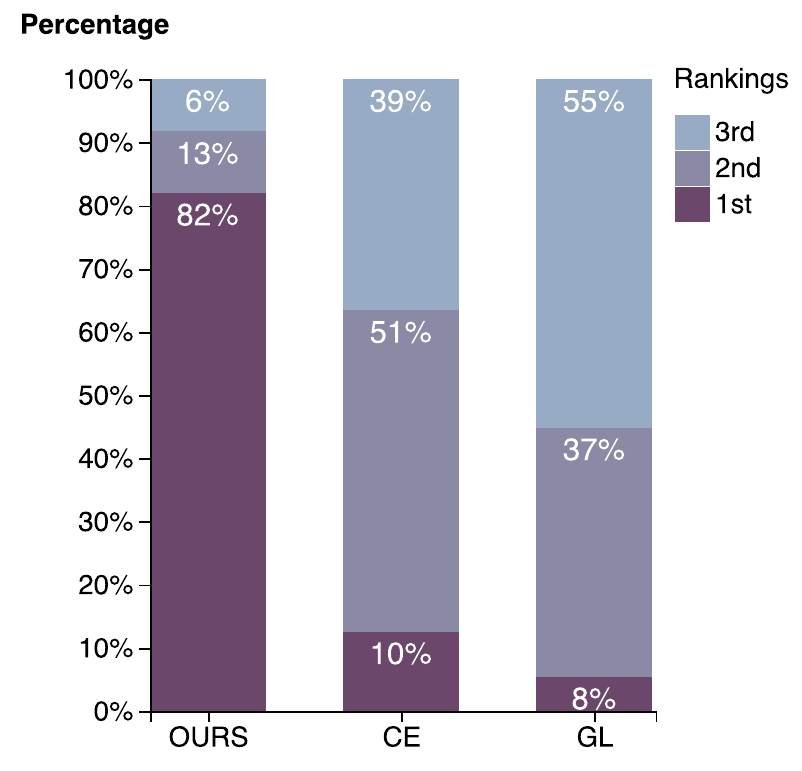} %
      \end{minipage}\hfill
	 \caption{The result of the user study to compare the naturalness of completion of three methods: ours, GL and CE. The three colors in each bar from bottom to top represent the percentage of each method being ranked first, second, and third. There are significantly more face images generated by our method being ranked the first than the other two approaches.}
     \label{fig:ranking}
\end{figure}%

\begin{table}
\caption{The pairwise t-test results of the user study. Our method was ranked first significantly more often than either CE or GL. There was no statistically significant difference in the likelihood of CE being ranked first versus GL.}
\label{tb:stat}
\begin{tabular}{|c |c| c| c|}
  \hline
  Method  & Vs. Ours & Vs. CE & Vs. GL \\
  \hline
  Ours    & - & t(31)=20.21 & t(31)=18.65 \\
          & - & \textbf{p<0.001}       & \textbf{p<0.001}       \\
  \hline
  CE      & t(31)=20.21 & - & t(31)=0.59 \\
          & \textbf{p<0.001}       & - & p=0.82 \\
  \hline
  GL      & t(31)=18.56  & t(31)=0.59  & -\\
          & \textbf{p<0.001}        & p=0.82  & -\\
  \hline
\end{tabular}
\end{table}

\subsubsection{Semantic Completion}
We first trained a high-resolution ($1024\times1024$) model with center masks (examples shown in Figure~\ref{fig:high}). In order to test whether our model actually learned high-level semantics and structures of faces, or simply ``remembered'' face examples, we designed two more challenging experiments.

In the \textit{first} experiment, training images were randomly flipped, rotated and covered with random masks. A new model was trained from scratch on this training set. The result (Figure~\ref{fig:high}) shows that our model was able to capture the anatomical structures of faces and generate content that is consistent with the holistic semantics.

In the \textit{second} experiment, we studied the ability of our model to infer high-resolution content from blurry contexts. A model pre-trained on clear contexts was used for this task. The testing images were down-sampled to $32\times32$ from original size with average pooling, and then up-sampled to $1024\times1024$ using bilinear interpolation. The result (Figure~\ref{fig:high}) demonstrates that our model also learned up-sampling information and was insensitive to blurry contexts.

\subsubsection{Attribute Controller}
Our network can be converted to a conditional version to control the attributes of generated images. Unlike the traditional image completion techniques that aim at reconstructing the missing parts so that they look similar to the original content, our goal is to complete faces with structurally meaningful content whose properties are controllable depending on the input attribute vectors, while making minimum changes to the context. In our experiment, two attributes (``Male'' and ``Smiling'') were chosen. This model was trained from scratch and the result was run at a $512\times512$ resolution (Figure~\ref{fig:attr}). The result shows that the attributes of synthesized images were controlled by our model explicitly. Additionally, the model learned to add detailed features (e.g. chin beard for the ``Male'' label) to the context to make the properties of synthesized faces more consistent with their attribute code, without affecting the holistic structures.

\subsubsection{Computation Time}
Existing CNN-based high-resolution in-painting approaches often need significant time to process an image. For instance, it took about 1 min for the model of Yang et al.~\cite{yang2016high} to fill in a $256\times256$ hole of a $512\times512$ image with a Titan X GPU. 

The advantage of our method is that our model, once trained, is able to complete a face image with a single forward pass, resulting in much higher efficiency. We tested the computation time of our model with a Titan Xp GPU by processing 3000 $1024\times1024$ images with $512\times512$ holes. The mean completion time of one image is 0.007 seconds with a standard deviation of 0.0005 seconds.

\begin{figure}
      \centering
      \begin{minipage}{0.47\textwidth}
          \centering
          \includegraphics[width=1.0\textwidth]{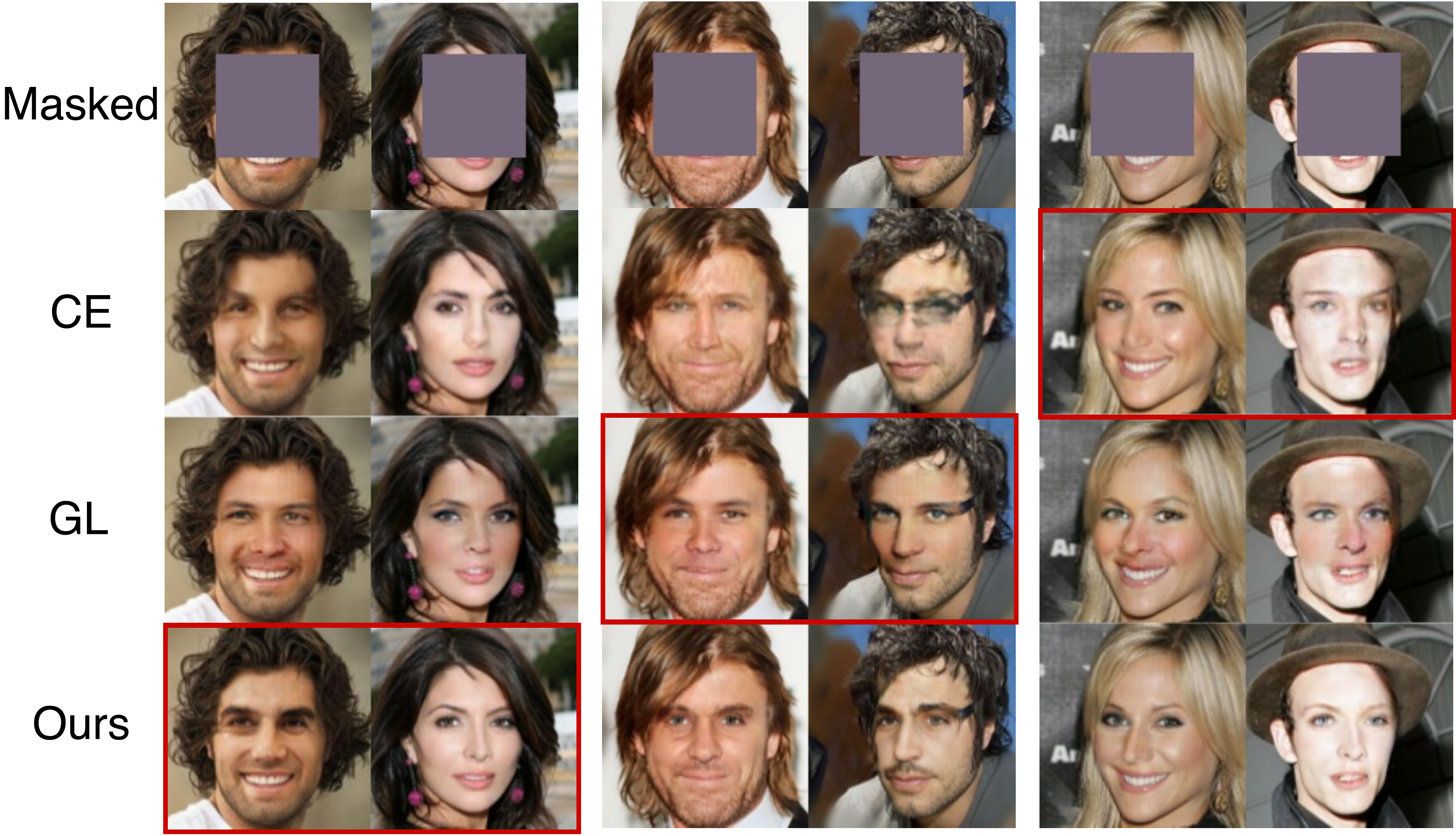} %
      \end{minipage}\hfill
	  \caption{Examples of images generated by different methods. The images being ranked first by users are annotated with red boxes.}
     \label{fig:samples}
\end{figure}%

\subsection{User Study}
We compared our method with two state-of-the-art CNN-based face completion approaches, CE and the Globally and Locally Consistent Method (GL)~\cite{iizuka2017globally}, with a pilot user study at $128\times128$ resolution with center masks. For GL, we used Poisson Blending~\cite{perez2003poisson} as post-processing. 32 subjects (21 male and 11 female participants, with ages from 22 to 36), including faculty and students, were volunteered to participate. 

For each trial, we randomly selected one synthesized image from each method and presented them on screen with permuted order. A user was asked to rank the three faces based on how realistic they looked (``1'' denoted the best while ``3'' denoted the worst). Each experiment started with a training session to help users become familiar with our user interface. The formal experiment consisted of 100 trials and there was no time limit. Most users finished the experiments within 20 minutes.

The result (Figure~\ref{fig:ranking}) shows that there were significantly more images generated by our method being favored by the viewers. Figure~\ref{fig:samples} shows some examples of face images produced by different methods. Overall, our approach generated sharper images with more details. However, sometimes users thought blurry images (e.g. female faces generated by CE) were more appealing. Without careful post processing, GL had a higher chance of producing content with inconsistent colors, for instance generating reddish faces while the contexts had pale skin. But, GL were better at object removal (e.g. removing glasses).

In order to confirm the intuition of our ranking results, we tested for statistical significance. To do this, we first collapsed each participant's rankings into a frequency list. For example, if a participant ranked our images first $77$ times, CE images $12$ times, and GL images $11$ times, the frequency list would be $77$, $12$, $11$. Given 32 participants, this resulted in 32 averages over 3200 samples.

Once frequency lists were built for all participants, the frequencies for each method (ours, CE, and GL) were again averaged over the 32 participants to produce a final list of averages from $n=32$ samples.

Next, we used Welch's analysis of variance (ANOVA)~\cite{mcdonald2009handbook} to test for statistically significant differences between the three ranking frequencies. We chose Welch's ANOVA rather than a standard ANOVA since we could not guarantee homogeneity of variance. Not surprisingly, results showed a significant difference in means for a standard $\alpha = 0.5$, with $F(2,29) = 213.6, p < 0.001$.

Given a significant ANOVA, we concluded by computing Games-Howell post-hoc pairwise tests (Table~\ref{tb:stat}) to see which methods' means differed significantly from one another (Games-Howell corrects for non-homogeneity of variance~\cite{ruxton2008time}). Results showed our method was significantly more likely to be ranked first versus both CE and GL. There was no statistically significant difference in the likelihood of CE being ranked first versus GL.

These results confirm that our method was ranked first significantly more often than either CE or GL.

\subsection{Limitations}
Though our method has low inference time, the training time is long due to the progressive growing of networks. In our experiment, it took about three weeks to train a $1024\times1024$ model on a Titan Xp GPU.

\begin{figure}
      \centering
      \begin{minipage}{0.47\textwidth}
          \centering
          \includegraphics[width=1.0\textwidth]{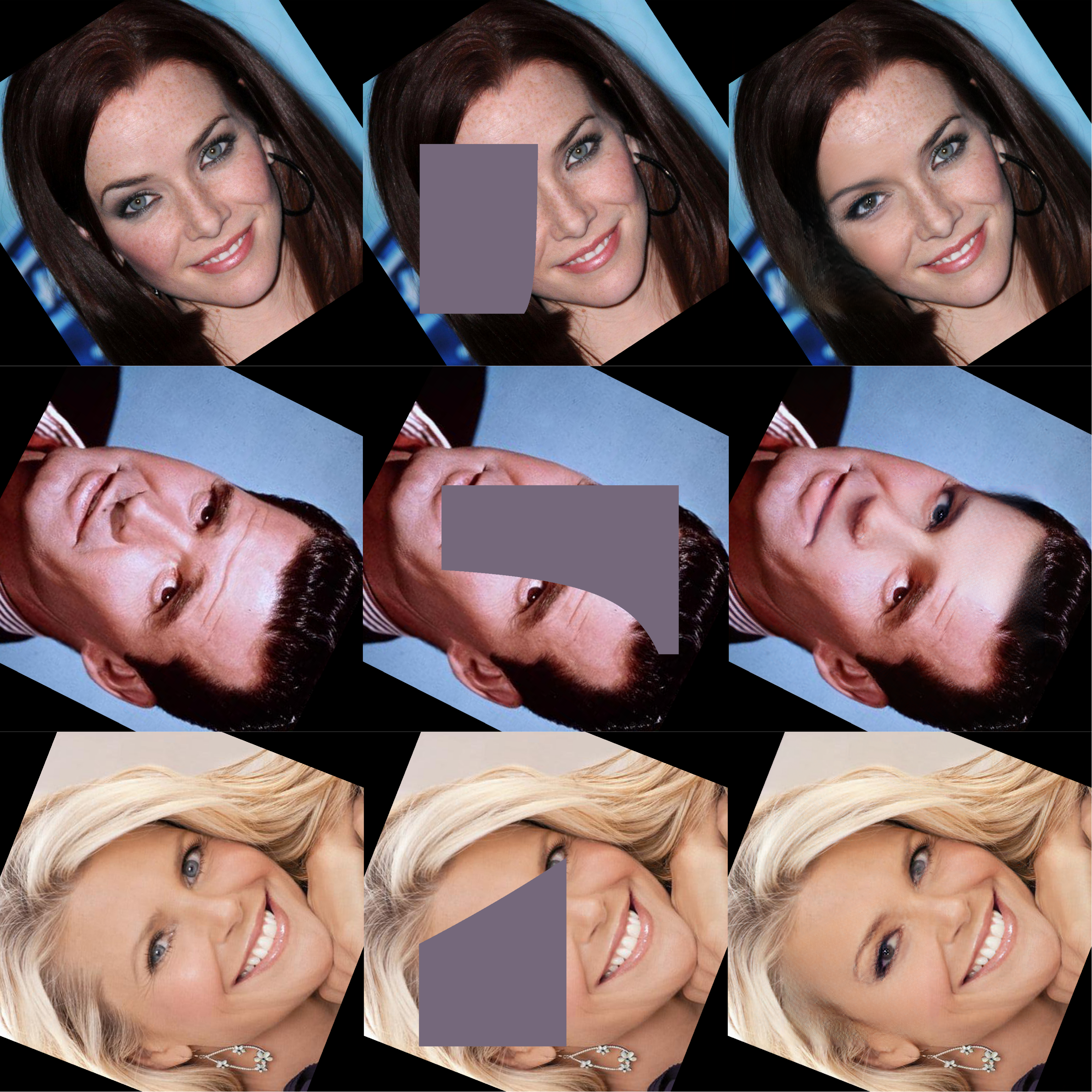} %
      \end{minipage}\hfill
	  \caption{Some failure cases of our approach. Our model tends to generate blurry images if the context has rich skin textures like freckles and wrinkles. It may also generate asymmetrical contents, for instance two eyes with different colors. The leftmost column are real images and the rightmost are synthesized faces by our approach.}
     \label{fig:fail}
\end{figure}

By carefully zooming in and inspecting our results, we find that our high-resolution model fails to learn low-level skin textures, such as furrows and sweat holes. Additionally, our model tends to generate blurry content while the context has abundant detailed textures (e.g. freckles). Moreover, occasionally, the model is unable to capture the symmetrical structure of faces (e.g. generating eyes with different colors). Some failure cases are shown in Figure~\ref{fig:fail}. These issues are left for future work.

\section{Conclusion}
We propose a deep learning approach for high-resolution face completion. Our model is trained progressively~\cite{karras2017progressive} and learns face structures from coarse to fine. By consolidating information across all scales, our model not only outperforms state-of-the-art methods by generating sharper images in low resolution, but is also able to synthesize faces in higher resolutions than existing techniques. A conditional version of our architecture allows users to control the properties of synthesized images explicitly with attribute vectors. Additionally, our architecture is designed in an end-to-end manner, in that it learns to generate completed faces directly with improved efficiency.

\begin{printonly}
\end{printonly}

\begin{screenonly}
\end{screenonly}


\bibliographystyle{ACM-Reference-Format}
\bibliography{sample-bibliography}